\documentclass[letterpaper, 10 pt, conference]{ieeeconf}  

\IEEEoverridecommandlockouts                              

\usepackage{color}
\usepackage{graphicx}
\usepackage{booktabs}
\usepackage{caption}
\usepackage{multirow}
\usepackage{multicol}
\usepackage{xcolor}
\usepackage{amssymb}
\usepackage{amsmath}
\usepackage{url}

\title{\LARGE \bf
PROFusion: Robust and Accurate Dense Reconstruction via Camera Pose Regression and Optimization
}

\author{
Siyan Dong$^{*}$,
Zijun Wang, 
Lulu Cai, 
Yi Ma, and Yanchao Yang \\
The University of Hong Kong
\thanks{$^{*}$ Email: siyan3d@hku.hk}
}

\begin{document}

\maketitle
\thispagestyle{empty}
\pagestyle{empty}

\begin{abstract}

Real-time dense scene reconstruction during unstable camera motions is crucial for robotics, yet current RGB-D SLAM systems fail when cameras experience large viewpoint changes, fast motions, or sudden shaking. Classical optimization-based methods deliver high accuracy but fail with poor initialization during large motions, while learning-based approaches provide robustness but lack sufficient accuracy for dense reconstruction. We address this challenge through a combination of learning-based initialization with optimization-based refinement. Our method employs a camera pose regression network to predict metric-aware relative poses from consecutive RGB-D frames, which serve as reliable starting points for a randomized optimization algorithm that further aligns depth images with the scene geometry. Extensive experiments demonstrate promising results: our approach outperforms the best competitor on challenging benchmarks, while maintaining comparable accuracy on stable motion sequences. 
The system operates in real-time, showcasing that combining simple and principled techniques can achieve both robustness for unstable motions and accuracy for dense reconstruction. 
Code released: 
\url{https://github.com/siyandong/PROFusion}.

\end{abstract}


\section{INTRODUCTION}

Real-time camera tracking and dense scene reconstruction are fundamental problems in robotics and computer vision. For autonomous robots, handling unstable camera motions is both challenging and critical. Current RGB-D SLAM (Simultaneous Localization and Mapping) systems perform well in controlled environments with smooth, typically slow camera movements. However, they struggle with the unstable motions encountered in practical applications like exploration or rescue missions - situations where robust camera pose estimation is essential for dense reconstruction.

Since the pioneering work of KinectFusion~\cite{newcombe2011kinectfusion,izadi2011kinectfusion}, the past decade has witnessed significant progress in RGB-D SLAM systems, particularly in the development of scene representations and camera pose estimation methods. Existing research has explored various representations such as volumetric~\cite{niessner2013real,chen2013scalable}, point-based~\cite{whelan2015elasticfusion,sandstrom2023point}, and neural representations~\cite{zhu2022nice,peng2024rtg,xin2024hero}. Building upon these representations, camera pose estimation is commonly performed through geometric optimization~\cite{hartley2003multiple,rusinkiewicz2001efficient} in combination with photometric losses~\cite{sucar2021imap,zhu2022nice}. While they can achieve high accuracy, they inherently require smooth and relatively slow camera motions - a limitation that restricts their widespread deployment in robotics. ROSEFusion~\cite{zhang2021rosefusion} recently introduced randomized optimization to better handle relatively fast camera motions, but it still struggles with rapid motions such as large in-place rotations. When cameras undergo large viewpoint changes, finding suitable initial poses becomes challenging. This makes the optimization process struggle to find the global optimum or even fail to converge.

\begin{figure}[tb!]
\centering
\includegraphics[width=1.0\linewidth]{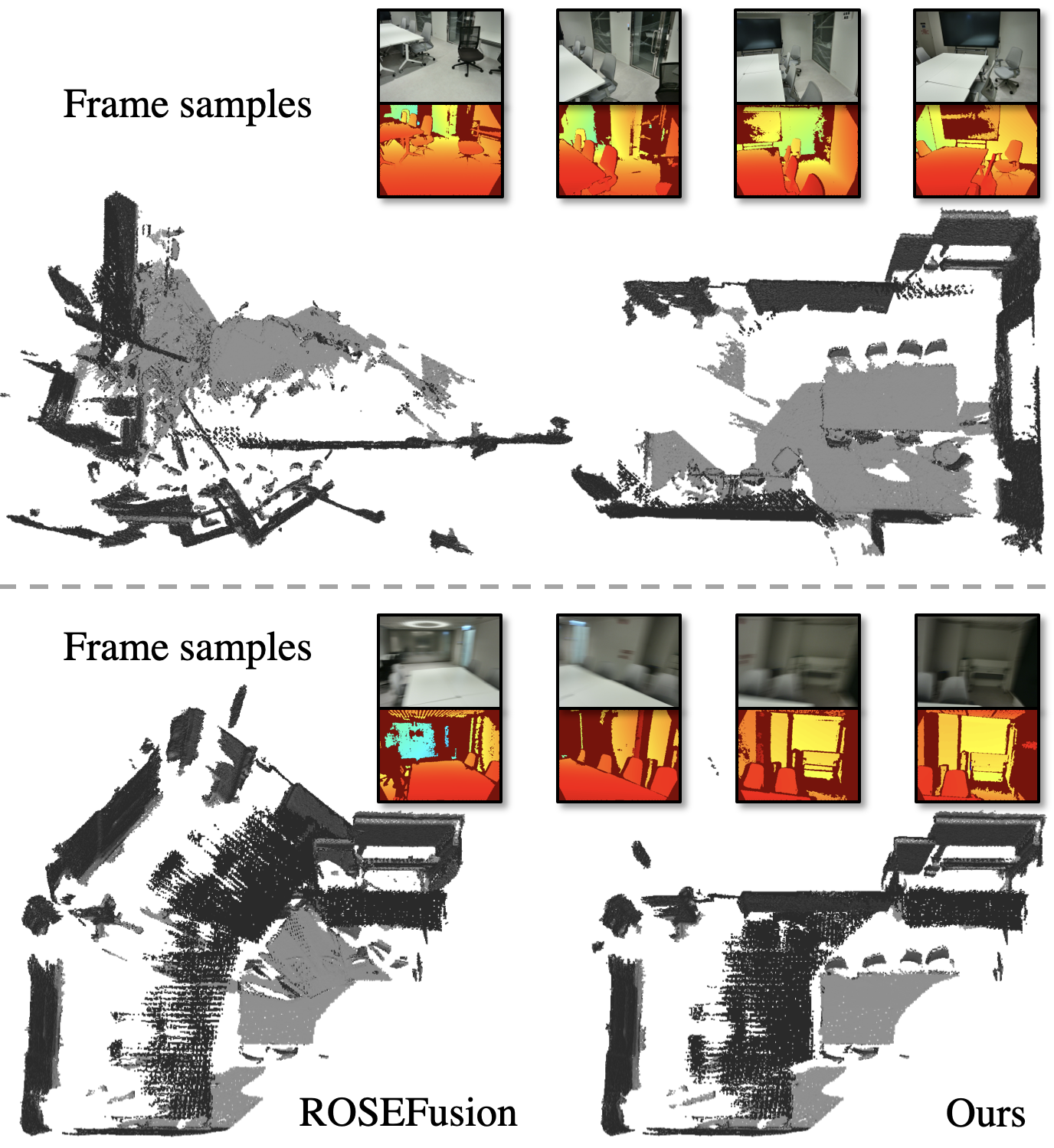}
\caption{
Dense scene reconstruction under challenging camera motions. Two representative sequences demonstrate failure cases where state-of-the-art methods like ROSEFusion~\cite{zhang2021rosefusion} (left) produce corrupted reconstructions due to unstable camera motions involving large translations and fast in-place rotations. Our approach (right), which combines camera pose regression and optimization, successfully reconstructs accurate scene layouts under challenging conditions, demonstrating superior robustness to camera motion instability. 
}
\label{fig:teaser}
\end{figure}

Unlike classical optimization methods, which can be sensitive to large motions, recent learning-based pose estimation approaches powered by large foundation models~\cite{wang2024dust3r,leroy2024grounding} offer much greater robustness. 
Thanks to scaling up the training on mixtures of diverse datasets, they demonstrate excellent generalization ability.
Among them, camera pose regression networks~\cite{dong2025reloc3r,wang2025vggt} achieve state-of-the-art success rates while maintaining fast inference times. 
However, existing works rarely take metric depth images as input, and the predicted poses are accurate only up to unknown scales. As a result, they often fall short of the tracking accuracy (e.g., millimeter-level error per frame) that classical optimization methods can achieve.

In this paper, we present a practical 3D reconstruction framework that combines the robustness of learning-based pose estimation with the accuracy of classical optimization. We integrate a metric-aware camera pose regression network with a randomized optimization algorithm to create an effective dense reconstruction system that achieves impressive results, despite its simplicity. We validate our approach through comprehensive experimental evaluations (e.g., real-world samples shown in Figure~\ref{fig:teaser}). 
Our key contributions can be summarized as follows:

\begin{itemize}

\item 
Our findings show that a camera pose regression network can reliably predict an initial coarse pose, which then serves as a starting point for further refinement using randomized optimization. 

\item 
Based on our findings, we have developed a real-time dense scene reconstruction system that provides robust and accurate camera tracking and scene reconstruction, regardless of camera motion stability.

\item 
Extensive experiments demonstrate that our system achieves accuracy on par with state-of-the-art dense reconstruction systems under stable camera motions, while offering substantially better robustness when handling camera shake, fast, and large motions.

\end{itemize} 
\section{RELATED WORKS}

\subsection{Dense scene reconstruction with RGB-D SLAM}
\label{sec:RGB-D SLAM}

KinectFusion~\cite{izadi2011kinectfusion,newcombe2011kinectfusion} pioneered real-time dense scene reconstruction with camera tracking using depth sensors. Subsequent research have introduced scalable scene representations~\cite{niessner2013real,chen2013scalable}, improved registration algorithms~\cite{rusinkiewicz2001efficient,halber2017fine}, and integrated color image feature matching~\cite{glocker2013real,dai2017bundlefusion,xin2024hero}. A core problem in RGB-D based reconstruction is accurate camera pose estimation. 
Registration errors can accumulate and cause drift in large scenes. Classical systems like InfiniTAM~\cite{kahler2015very}, ElasticFusion~\cite{whelan2015elasticfusion}, and BundleFusion~\cite{dai2017bundlefusion} address this problem by integrating loop closure and bundle adjustment to achieve globally consistent reconstruction. 

Recent SLAM systems have adopted neural scene representations~\cite{zhu2022nice,peng2024rtg,xin2024hero}, enabling photo-realistic novel view synthesis.
The photometric losses used in these systems can refine camera poses and reduce error, but they may be sensitive to noise like motion blur or exposure changes during fast motions. As a result, their robustness still depends on bundle adjustment or additional backend systems~\cite{mur2017orb}.

There are also non-rigid~\cite{newcombe2015dynamicfusion} or semantic reconstruction~\cite{mccormac2017semanticfusion} approaches that are beyond our focus in this paper. 

\subsection{Randomized optimization for fast camera motions}
A key limitation of the approaches reviewed in Section~\ref{sec:RGB-D SLAM} is their requirement for smooth and typically very slow camera motions. This poses challenges for robotic applications, such as exploration and rescue scenarios, where cameras often shake on rugged terrain and move fast. Such conditions cause motion blur and significant viewpoint changes between frames, making camera tracking difficult.
ROSEFusion~\cite{zhang2021rosefusion} uses randomized optimization to address nonlinearity from viewpoint changes. The intuition is to apply a wide range of random search to find solutions. However, as relative poses increase in magnitude, the required search space grows proportionally, making the method impractical for large motions. A typical failure case occurs with fast in-place rotation - a common motion in robots.

More recent works like MIPS-Fusion~\cite{tang2023mips} and RemixFusion~\cite{lan2025remixfusion} combine randomized optimization with neural representations. While achieving promising scalability and rendering quality, these approaches sacrifice robustness. In this paper, we focus on achieving robustness and accuracy for unstable camera movements - particularly the large viewpoint changes caused by fast motion, low frame rates, or signal losses. Our experiments indicate that ROSEFusion~\cite{zhang2021rosefusion} remains the strongest competitor in this field.

\subsection{Large 3D foundation models}

DUSt3R~\cite{wang2024dust3r} was the first to leverage a large mixture of public datasets to train a large geometric foundation model, demonstrating that high-quality 3D reconstruction and remarkable generalization capability can emerge through the scaling laws. 
It triggered a surge of neural networks for geometric reasoning, quickly producing numerous research outcomes in areas such as keypoint matching~\cite{leroy2024grounding}, camera pose estimation~\cite{dong2025reloc3r}, and multi-view 3D reconstruction~\cite{wang20253d,tang2025mv,yang2025fast3r,elflein2025light3r,wang2025continuous,liu2025slam3r,murai2025mast3r,wang2025vggt,keetha2025mapanything,cheng2025outdoor}. 

Among these works, SLAM3R~\cite{liu2025slam3r} is the only one compatible with point cloud input, and Reloc3r~\cite{dong2025reloc3r} provides the simplest yet most efficient approach for camera pose estimation. However, their predictions are only accurate up to unknown scale factors. 
This paper draws inspiration from point cloud embedding~\cite{liu2025slam3r} but utilizes metric point clouds obtained from depth scanning. By integrating this approach with camera pose regression~\cite{dong2025reloc3r}, we create a simple yet robust metric camera pose estimation network. 
\begin{figure*}[tbh!]
\centering
\includegraphics[width=1.0\linewidth]{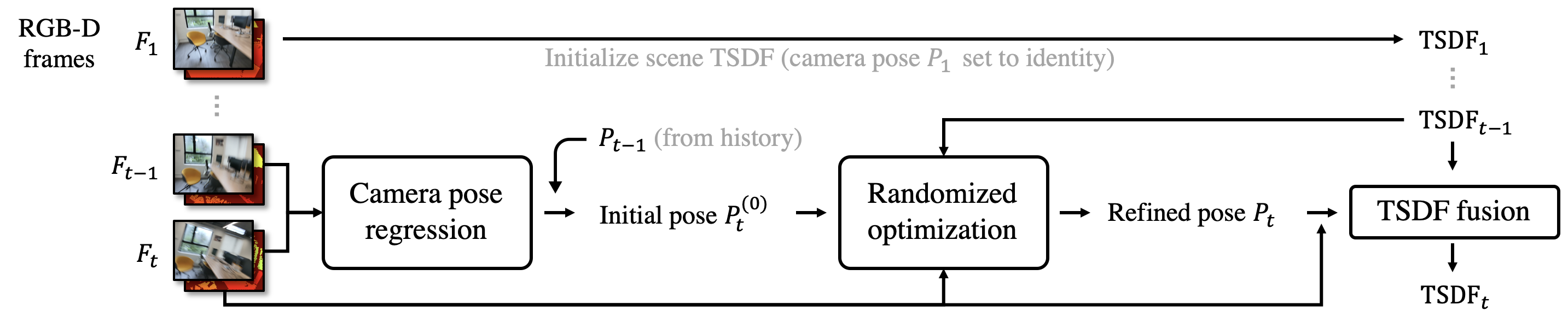}
\caption{
System overview. We use the first frame (set to identity pose) to initialize the scene represented by TSDF grids. The following frames are incrementally fused to the scene through a two-step process: first, a coarse registration with the previous frame via camera pose regression, and second, a fine-grained alignment to the TSDF via a randomized optimization algorithm. The aligned frames then update the TSDF values by modifying known grids and filling in new ones. Through this process, both camera motion and scene geometry are progressively reconstructed. 
}
\label{fig:pipeline}
\end{figure*}

\begin{figure}[tbh!]
\centering
\includegraphics[width=1.0\linewidth]{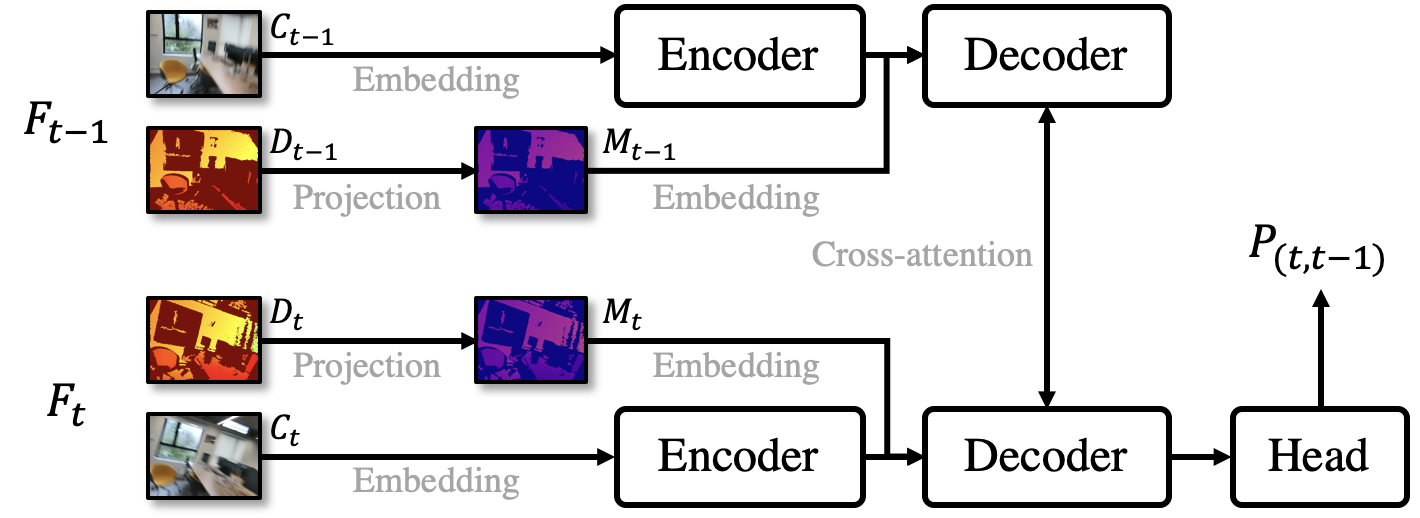}
\caption{Network architecture. It takes a pair of consecutive RGB-D frames as input and outputs the relative camera pose that aligns the second frame to the first one. The color and depth (converted to metric point clouds) images are divided into tokens and fed into a Transformer backbone with a pose regression head to infer the relative transformation matrix. } 
\label{fig:network}
\end{figure}

\begin{figure}[tbh!]
\centering
\includegraphics[width=1.0\linewidth]{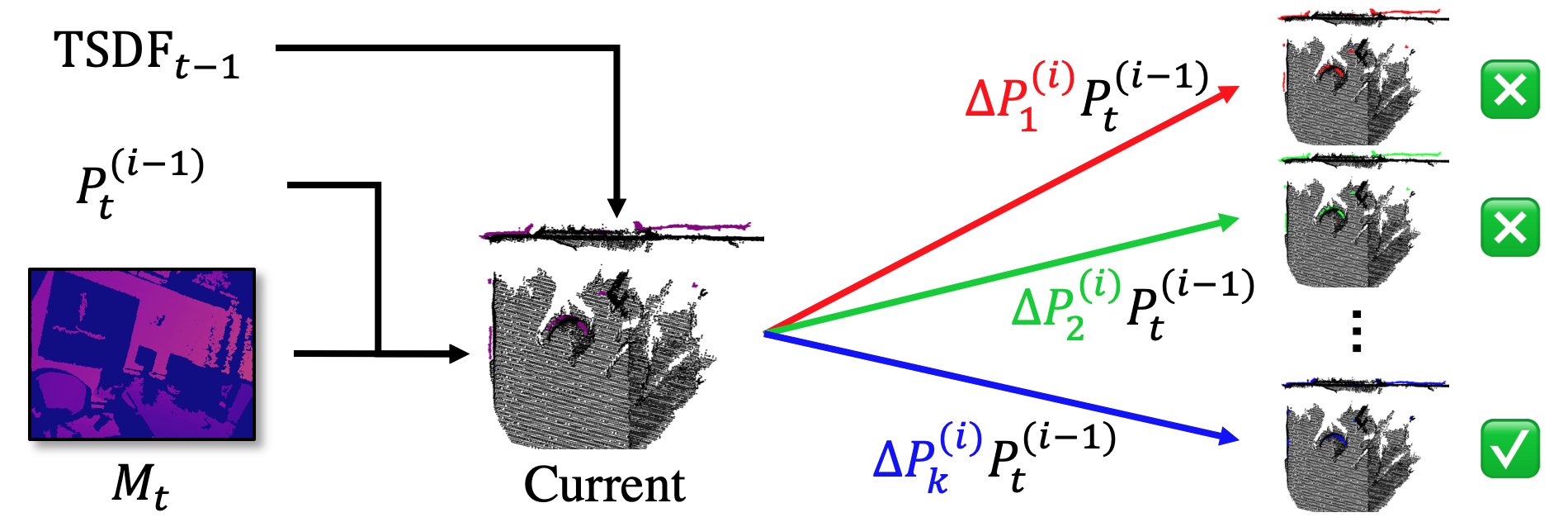}
\caption{
Illustration of the camera pose searching process in iteration $i$ in our randomized optimization. We multiply a set of delta poses $\{ \Delta P_k^{(i)} \}_{k=1}^K$ to the current pose $P_t^{(i-1)}$ and evaluate their fitness to TSDF$_{t-1}$. Delta poses with better alignment are collected in an advantage set to update the current pose and the search size for the next iteration. 
}
\label{fig:ro}
\end{figure}

\section{METHOD}

\noindent
\textbf{Problem setting.}
Our system takes an RGB-D video as input, consisting of consecutive frames $\{F_t=(C_t, D_t)\}_{t=1}^T$ captured by a moving camera in a static scene. We assume aligned color $C_t \in \mathbb{R}^{H \times W \times 3}$ and depth $D_t \in \mathbb{R}^{H \times W \times 3}$ images for each frame, with known camera intrinsic parameters. Therefore, each depth image $D_t$ can be projected to a metric point cloud $M_t \in \mathbb{R}^{H \times W \times 3}$. 
Our goal is to recover the camera poses $\{P_t = [ \mathbf{R}_t | \mathbf{t}_t ] \}_{t=1}^T$ (where $\mathbf{R} \in SO(3)$ is the rotation matrix and $\mathbf{t} \in \mathbb{R}^3$ represents the translation in the world coordinate system) of the frames and reconstruct the dense 3D geometry of the captured scene. To simplify notation, we denote the camera pose in the world coordinate system as $P_i$ and relative pose from frame $F_i$ to $F_j$ as $P_{(i,j)}$.

\noindent
\textbf{Method overview.}
Figure~\ref{fig:pipeline} illustrates the pipeline of our system. Following KinectFusion~\cite{newcombe2011kinectfusion,izadi2011kinectfusion}, we represent the scene using truncated signed distance function (TSDF)~\cite{curless1996volumetric}. This is implemented with a tensor that records the distance value in each 3D grid, denoting the distance to its closest surface. The first input frame is set to the identity pose, and its depth image initializes the TSDF values.
The key feature of our system is robust and accurate camera pose estimation, which combines learning-based initialization with optimization-based refinement.
Starting from the second frame, each frame $F_t$ is paired with its previous frame $F_{t-1}$ and fed into a neural network to estimate an initial camera pose, denoted by $P_t^{(0)}$. This pose serves as the starting point for aligning the depth image $D_t$ to the scene TSDF$_{t-1}$ through randomized optimization. 
The optimized pose, denoted by $P_t$, is recorded for future pose estimation, and the scene is updated to TSDF$_{t}$ after fusing $F_t$. 
The fusion process remains the same as KinectFusion~\cite{newcombe2011kinectfusion,izadi2011kinectfusion}.
In the following sections, we first present our camera pose regression network in Section~\ref{sec:pr}. Next, we introduce our adapted randomized optimization algorithm in Section~\ref{sec:ro}. Finally, we report implementation details in Section~\ref{sec:implementation}.

\subsection{Camera Tracking Initialization via Pose Regression}
\label{sec:pr}

Our pose regression network takes a pair of consecutive RGB-D frames $(F_{t-1}, F_t)$ as input and outputs the relative camera pose $P_{(t,t-1)}$. The absolute pose $P_t^{(0)}$ in the world coordinate system is then calculated by multiplying this relative pose with the previous pose estimation $P_{t-1}$, denoted as: $P_t^{(0)} = P_{t-1} P_{(t,t-1)}$. 

Our network architecture draws inspiration from recent foundation models~\cite{wang2024dust3r,liu2025slam3r,dong2025reloc3r}. 
For the sake of simplicity, we adopt DUSt3R~\cite{wang2024dust3r}'s backbone with minimal modifications. The network architecture is illustrated in Figure~\ref{fig:network}. It consists of a two-branch Vision Transformer (ViT)~\cite{dosovitskiy2020image}. 
The ViT encoder extracts features from each image, and the decoder exchanges information between branches. 
Finally, a regression head outputs a matrix as the relative pose. 

\noindent
\textbf{Frame embedding.} 
Each color image $C_i$ undergoes the default ViT patch embedding process to produce color tokens $\mathbf{C}_i$. Each depth image is first projected to a metric point cloud $M_i$ using known camera intrinsics. This point cloud then undergoes the same patch embedding process (without normalization) to extract metric-aware geometry tokens $\mathbf{M}_i$.

\noindent
\textbf{ViT backbone.} 
The color tokens are fed into $m$ ViT encoder blocks for feature extraction: $\mathbf{C}_i'= $ Encoder($\mathbf{C}_i$). To preserve the metric information, we do not apply an encoding process to geometry tokens $\mathbf{M}_i$. These tokens are directly added to the encoded color tokens to form feature tokens $\mathbf{Z}_i=\mathbf{C}_i'+\mathbf{M}_i$ for the following decoding process.
Unlike the encoder blocks that operate on tokens in each frame separately, the $n$ decoder blocks incorporate cross-attention layers to reason the spatial transformation between the two sets of tokens. We denote the decoded tokens as: $\mathbf{Z}_t'=$ Decoder($\mathbf{Z}_t$, $\mathbf{Z}_{t-1}$).

\noindent
\textbf{Pose regression.} 
Our pose head remains similar to Reloc3r~\cite{dong2025reloc3r}'s regression head. The architecture is the same; the only difference is that our translation is set to be metric, while theirs only predicts direction.
Our pose regression process can be denoted as: $P_{(t,t-1)} = $ Head($\mathbf{Z}_t'$). 

\noindent
\textbf{Supervision.}
We train our network using supervision on metric relative poses. The loss function minimizes the angular error~\cite{dong2025reloc3r} in rotation and the distance in translation: 
\begin{align*} 
& \mathcal{L} = \ell_\mathbf{R} + \ell_\mathbf{t}, \\
& \ell_\mathbf{R} = \text{arccos} ( \frac{\text{tr}(\bar{\mathbf{R}}^{-1} \mathbf{R}) - 1}{2} ), \quad \text{and} \quad  
\ell_\mathbf{t} = \text{norm} (\bar{\mathbf{t}} - {\mathbf{t}}).
\end{align*}
Here, $\bar{\mathbf{R}}$ and $\bar{\mathbf{t}}$ represent the ground truth rotation and translation of the relative pose, respectively. 
Our network is trained on a mixture of public datasets and generalizes well. Details are reported in Section~\ref{sec:implementation}.

\subsection{Pose Refinement via Randomized Optimization} 
\label{sec:ro}

Our randomized optimization algorithm takes the current scene TSDF$_{t-1}$, point cloud $M_t$, and the initial pose $P_t^{(0)}$ as input, then iteratively searches for incremental relative poses (which we call delta poses) that can better align $M_t$ to TSDF$_{t-1}$. 
Note that we do not use color images in this process. Under unstable camera motions, color images can become blurred, thereby reducing accuracy. 
While depth images can be incomplete, their valid pixels remain accurate. 
Building on this property, our algorithm draws inspiration from ROSEFusion~\cite{zhang2021rosefusion}, but is simplified as detailed below.

Let's denote the search size as $s=(\omega,v)$, representing $\omega$ degree and $v$ cm searching range. Our algorithm iteratively updates the current pose $P_t^{(i-1)}$ and search size $s^{(i)}$.  
In each iteration $i$, as illustrated in Figure~\ref{fig:ro}, 
we have a current pose $P_t^{(i-1)}$ from previous iterations. 
We uniformly sample a set of delta poses $\{ \Delta P_k^{(i)} \}_{k=1}^K$ according to $s^{(i)}$. 
Next, we multiply each delta pose $\Delta P_k^{(i)}$ to $P_t^{(i-1)}$ to formulate a global transformation, $\Delta P_k^{(i)} P_t^{(i-1)}$, that aligns $M_t$ to TSDF$_{t-1}$. 
We then evaluate if this transformation actually improves the alignment. We quantify this by measuring the geometric consistency error between the transformed point cloud and the current scene representation:

\begin{equation*}
\mathcal{E}(\Delta P_k^{(i)}P_t^{(i-1)}) = 
\frac{1}{|X_k|} \sum_{x\in X_k} \frac{|\text{TSDF}_{t-1}(\Delta P_k^{(i)} P_t^{(i-1)} x) |}{\tau}.
\end{equation*}
Here, $\tau$ represents the truncated distance, $x$ denotes a 3D points from $M_t$, and $X_k$ represents the subset of points that fall in TSDF$_{t-1}$'s valid grids (i.e., grids observed by history frames and within distance $\tau$) when applying the transformation $\Delta P_k^{(i)} P_t^{(i-1)} $. 
$|\text{TSDF}_{t-1}(\cdot)|$ represents the absolute distance value queried by the input point. This is normalized to the range [0, 1] by multiplying $1 / \tau$. $|X_k|$ represents the number of points in $X_k$. 
Delta poses with lower errors (i.e., $\mathcal{E}(\Delta P_k^{(i)}P_t^{(i-1)})<\mathcal{E}(P_t^{(i-1)})$) are collected in an advantage set. 
We compute an average (rotation in $SO(3)$ and translation in $\mathbb{R}^3$) delta pose within this set, denoted by $\hat{\Delta P^{(i)}}$. 
The current pose is updated by $P_t^{(i)} = \hat{\Delta P^{(i)}} P_t^{(i-1)}$, and the search size is updated by $s^{(i+1)} = \beta s^{(i)} + (1-\beta) \mathcal{E}( P_t^{(i)} ) \ s^{(i)}$. Here, $\beta=0.1$ represents the momentum between iterations. 
As a result, the search size converges during iterations.

\section{EXPERIMENT}

\subsection{Implementation details}
\label{sec:implementation}

\noindent
\textbf{Camera pose regression network.}
This module is implemented in Python with PyTorch. 
The images input to our network are resized and center-cropped to a fixed resolution of $224 \times 224$ for efficiency. Following DUSt3R~\cite{wang2024dust3r}, we employ a 24-block encoder and a 12-block decoder.  
The network weights are initialized from SLAM3R~\cite{liu2025slam3r}'s L2W model and then trained with a mixture of public datasets featuring indoor RGB-D scanning with known camera poses. The training data includes ScanNet++~\cite{yeshwanth2023scannet++}, Aria Synthetic Environments~\cite{avetisyan2024scenescript}, and few sequences from 7 Scenes~\cite{shotton2013scene} and Replica~\cite{straub2019replica} to ensure diversity. 
This resulted in around 2 million RGB-D image pairs, with relative camera poses ranging from 0-180 degrees and 0-5 meters.
It's important to note that all test scenes were excluded from the training data, and our network was trained only once to handle all tests. 
The network demonstrates strong generalization capability across various benchmarks and real-world applications. 

\noindent
\textbf{Randomized optimization algorithm.}
This module is implemented in C++ with CUDA. 
The initial search size $s^{(1)}$ is set to (10 degree, 10 cm) by default. 
Following ROSEFusion~\cite{zhang2021rosefusion}, we alternate the number of delta poses (1024, 3072, and 10240) across iterations. 
Similarly, we down-sample the input point cloud for efficiency, and alternate the sampling rates (1/8, 1/16, and 1/32). 
While we also set a maximum of 20 iterations, our optimization typically converges within just a few iterations.

\subsection{Comparisons}

\begin{table*}[tbh!]

\caption{Camera tracking accuracy evaluation using ATE-RMSE (cm) on FastCaMo-Synth (raw) benchmark.}

\centering

\resizebox{1.0\textwidth}{!}{


\begin{tabular}{l|cccccccccc|c}
\toprule
Method & Apart.\_1 & Apart.\_2 & Frl\_a.\_2 & Hotel\_0 & Office\_0 & Office\_1 & Office\_2 & Office\_3 & Room\_0 & Room\_1 & Avg. \\
\toprule

ElasticFusion~\cite{whelan2015elasticfusion} & 23.2 & 56.8 & 12.6 & 36.6 & 19.2 & 21.9 & - & 80.0 & - & 43.6 & - \\

BundleFusion~\cite{dai2017bundlefusion} & {3.0} & {{1.1}} & 2.8 & 54.4 & {{0.8}} & - & - & - & {\underline{1.5}} & - & - \\

ROSEFusion~\cite{zhang2021rosefusion} & {\underline{0.7}} & {\textbf{0.6}} & {\textbf{0.5}} & {\underline{0.8}} & {\textbf{0.5}} & {\underline{0.8}} & 8.3 & {\underline{10.1}} & 2.1 & {\underline{2.0}} & {\underline{2.6}} \\

{NICE-SLAM~\cite{zhu2022nice}} & - & 36.7 & 15.4 & 4.2 & 8.4 & 13.7 & 14.6 & 14.3 & - & 29.7 & - \\

MIPS-Fusion~\cite{tang2023mips} & 7.0 & 1.5 & {{1.9}} & 4.8 & 3.6 & 5.6 & {\underline{7.4}} & 17.4 & 4.4 & 5.1 & 5.9 \\

{HERO-SLAM~\cite{xin2024hero}} & 3.7 & {\underline{0.7}} & {\underline{0.7}} & 1.7 & {\underline{0.7}} & 1.1 & 13.4 & 24.6 & 7.6 & {\textbf{0.5}} & 5.5 \\

\midrule

{Ours} & {\textbf{0.5}} & {\textbf{0.6}} & {\textbf{0.5}} & {\textbf{0.7}} & {\textbf{0.5}} & {\textbf{0.5}} & {\textbf{1.8}} & {\textbf{1.0}} & {\textbf{0.4}} & {\textbf{0.5}} & {\textbf{0.7}} \\

\bottomrule
\end{tabular}

}

\label{tab:fastcamo_raw}
\end{table*}

\begin{table*}[tbh!]

\caption{Camera tracking accuracy evaluation using ATE-RMSE (cm) on FastCaMo-Synth (noise) benchmark.}

\centering

\resizebox{1.0\textwidth}{!}{


\begin{tabular}{l|cccccccccc|c}
\toprule
Method & Apart.\_1 & Apart.\_2 & Frl\_a.\_2 & Hotel\_0 & Office\_0 & Office\_1 & Office\_2 & Office\_3 & Room\_0 & Room\_1 & Avg. \\
\toprule

ElasticFusion~\cite{whelan2015elasticfusion} & 40.9 & 40.7 & 43.8 & 43.8 & 22.3 & 2.3 & - & 94.3 & - & 31.0 & - \\

BundleFusion~\cite{dai2017bundlefusion} & 4.6 & 2.2 & 83.6 & 65.2 & 2.7 & 17.3 & - & - & - & - & - \\

ROSEFusion~\cite{zhang2021rosefusion} & {\underline{1.5}} & {\textbf{0.9}} & 3.0 & {\underline{1.6}} & {\underline{0.7}} & {1.8} & {\underline{3.6}} & 9.4 & {\underline{2.9}} & 3.8 & {\underline{2.9}} \\ 

{NICE-SLAM~\cite{zhu2022nice}} & - & 20.2 & 24.8 & 11.8 & 29.3 & - & 16.4 & 29.8 & - & 24.9 & - \\

MIPS-Fusion~\cite{tang2023mips} & 6.6 & 3.1 & {\textbf{2.6}} & 5.2 & 7.6 & 17.4 & 24.9 & {\underline{6.0}} & 4.4 & {\underline{3.6}} & 8.1 \\

{HERO-SLAM~\cite{xin2024hero}} & 3.7 & {\underline{1.2}} & 7.5 & {\underline{1.6}} & 1.2 & {\underline{1.7}} & 14.7 & 23.7 & 11.7 & {\textbf{1.3}} & 6.8 \\ 

\midrule

{Ours} & {\textbf{1.2}} & {\textbf{0.9}} & {\underline{2.9}} & {\textbf{1.5}} & {\textbf{0.5}} & {\textbf{1.1}} & {\textbf{1.8}} & {\textbf{1.9}} & {\textbf{1.7}} & {\textbf{1.3}} & {\textbf{1.5}} \\ 

\bottomrule
\end{tabular}

}

\label{tab:fastcamo}
\end{table*}

\begin{table}[tbh!]

\caption{Camera tracking accuracy evaluation using ATE-RMSE (cm) on stable motions from TUM RGB-D dataset.
The methods marked with * represent only single-frame camera tracking w/o bundle adjustment or loop optimization. }

\centering

\resizebox{0.46\textwidth}{!}{


\begin{tabular}{l|ccc|c}
\toprule
Method & fr1\_desk & fr2\_xyz & fr3\_office & Avg. \\
\toprule

ElasticFusion~\cite{whelan2015elasticfusion} & {\underline{2.0}} & {\textbf{1.1}} & {\textbf{1.7}} & {\textbf{1.6}} \\

BundleFusion~\cite{dai2017bundlefusion} & {\textbf{1.6}} & {\textbf{1.1}} & {\underline{2.2}} & {\textbf{1.6}} \\ 

ROSEFusion~\cite{zhang2021rosefusion} * & 2.3 & 3.3 & 3.9 & 3.1 \\ 

NICE-SLAM~\cite{zhu2022nice} & 2.7 & {{1.8}} & 3.0 & 2.5 \\

MIPS-Fusion~\cite{tang2023mips} & 3.0 & {\underline{1.4}} & 4.6 & 3.0 \\

HERO-SLAM~\cite{xin2024hero} & 2.5 & 2.1 & 2.7 & {{2.4}} \\ 

\midrule

{Ours} * & {\underline{2.0}} & 2.3 & {{2.5}} & {\underline{2.3}} \\

\bottomrule
\end{tabular}

}

\label{tab:tum}
\end{table}

\begin{table}[tbh!]

\caption{Camera tracking accuracy evaluation using ATE-RMSE (cm) on camera shaking scenes from ETH3D dataset.}

\centering

\resizebox{0.43\textwidth}{!}{


\begin{tabular}{l|ccc|c}
\toprule
Method & camera\_shake\_1 & 2 & 3 & Avg. \\
\toprule

ElasticFusion~\cite{whelan2015elasticfusion} & 8.4 & - & - & - \\

BundleFusion~\cite{dai2017bundlefusion} & 5.2 & 3.5 & - & - \\ 

ROSEFusion~\cite{zhang2021rosefusion} & {\underline{0.9}} & {1.8} & {\underline{5.0}} & {\underline{2.6}} \\

{NICE-SLAM~\cite{zhu2022nice}} & 1.0 & - & - & - \\ 

{MIPS-Fusion~\cite{tang2023mips}} & 4.6 & - & 7.7 & - \\ 

{HERO-SLAM~\cite{xin2024hero}} & {\textbf{0.7}} & {\underline{1.5}} & - & - \\ 

\midrule

{Ours} & {\textbf{0.7}} & {\textbf{1.2}} & {\textbf{3.6}} & {\textbf{1.8}} \\ 

\bottomrule
\end{tabular}

}

\label{tab:eth3d}
\end{table}

\begin{table}[tbh!]

\caption{
Evaluation of reconstruction completeness and accuracy on FastCaMo-Real benchmark.
}

\centering

\resizebox{0.5\textwidth}{!}{


\begin{tabular}{l|cccc|c}
\toprule
Method & Apart.\_1 & Apart.\_2 & Gym & Lab & Avg. \\
\toprule

\multirow{2}{*}{ROSEFusion~\cite{zhang2021rosefusion}} & {\underline{84.4\%}} & {\textbf{84.4\%}} & {\underline{43.0\%}} & {\textbf{84.3\%}} & {\underline{74.0\%}} \\

& 4.4cm & {\underline{3.1cm}} & {\underline{4.9cm}} & {\textbf{2.9cm}} & {\underline{3.8cm}} \\

\midrule

\multirow{2}{*}{{MIPS-Fusion~\cite{tang2023mips}}} & 76.2\% & 70.6\% & - & 71.8\% & - \\

& {\underline{4.2cm}} & 4.2cm & - & 3.4cm & - \\

\midrule

\multirow{2}{*}{{HERO-SLAM~\cite{xin2024hero}}} & 73.2\% & 71.1\% & - & 50.5\% & - \\

& 5.1cm & 4.7cm & - & 4.6cm & - \\

\midrule

\multirow{2}{*}{{Ours}} & {\textbf{89.7\%}} & {\underline{84.0\%}} & {\textbf{56.0\%}} & {\underline{84.2\%}} & {\textbf{78.5\%}} \\ 

& {\textbf{3.3cm}} & {\textbf{3.0cm}} & {\textbf{4.5cm}} & {\underline{3.0cm}} & {\textbf{3.5cm}} \\

\bottomrule
\end{tabular}

}

\label{tab:fastcamo_real}
\end{table}

\begin{table}[tbh!]

\caption{Evaluation of reconstruction completeness and accuracy on FastCaMo-Real benchmark. To mimic unstable motion, we uniformly drop 20\% of frames for each sequence.}

\centering

\resizebox{0.5\textwidth}{!}{


\begin{tabular}{l|cccc|c}
\toprule
Method & Lou.\_1 & Lou.\_2 & Meet. & Office & Avg. \\
\toprule

\multirow{2}{*}{ROSEFusion~\cite{zhang2021rosefusion}} & 76.8\% & 80.2\% & \textbf{76.8\%} & 44.7\% & 69.6\% \\

& 2.2cm & 4.1cm & 3.8cm & 3.3cm & 3.4cm \\

\midrule

\multirow{2}{*}{{Ours}} & \textbf{77.8\%} & \textbf{94.3\%} & {75.7\%} & \textbf{54.9\%} & \textbf{75.7\%} \\

& \textbf{2.1cm} & \textbf{2.9cm} & \textbf{3.7cm} & \textbf{2.1cm} & \textbf{2.7cm} \\

\bottomrule
\end{tabular}

}

\label{tab:fastcamo_real2}
\end{table}

For our comparison, we selected six representative dense scene reconstruction systems. 
The systems include three state-of-the-art classical dense fusion approaches (ElasticFusion~\cite{whelan2015elasticfusion}, BundleFusion~\cite{dai2017bundlefusion}, and ROSEFusion\cite{zhang2021rosefusion}) and three neural SLAM systems (NICE-SLAM~\cite{zhu2022nice}, MIPS-Fusion~\cite{tang2023mips}, and HERO-SLAM~\cite{xin2024hero}). 
We test these methods on classical benchmarks with stable camera motions (TUM RGB-D~\cite{sturm2012benchmark}), but primarily focus on benchmarks with unstable motions: camera shaking scenarios from ETH3D~\cite{schops2019bad} and fast motion from FastCaMo benchmarks~\cite{zhang2021rosefusion}, which include both synthetic and real-world RGB-D scans. 
The results are produced using publicly available code and papers. In the tables, a minus symbol (``-'') indicates cases where the method failed to reconstruct at least 40\% of the sequence or produced a completely incorrect trajectory misaligned with ground-truth.
Additionally, we conduct real-world applications with an RGB-D camera, ORBBEC Femto Bolt.

\noindent
\textbf{TUM RGB-D~\cite{sturm2012benchmark}.} 
We report camera tracking accuracy in Table~\ref{tab:tum}. The camera motions in these sequences are smooth and slow. Our method achieves comparable accuracy to the best-performing systems, ElasticFusion~\cite{whelan2015elasticfusion} and BundleFusion~\cite{dai2017bundlefusion}, which use global optimization with loop closures to correct accumulated drift. Note that NICE-SLAM~\cite{zhu2022nice}, HERO-SLAM~\cite{xin2024hero}, and MIPS-Fusion~\cite{tang2023mips} also employ bundle adjustment to jointly optimize camera poses across keyframes. Despite using only single-frame tracking, our system achieves slightly lower tracking errors. This validates the accuracy of our approach.

\noindent
\textbf{ETH3D~\cite{schops2019bad}.} 
Our method demonstrates its advantage when dealing with unstable camera motions. To evaluate this, we conduct a comparison on three challenging sequences from the ETH3D benchmark that feature camera shaking. Camera shake results in sudden speed changes and blurred color images. This variation in motion between frames can be significant, with some frames experiencing large viewpoint changes. The results are reported in Table~\ref{tab:eth3d}. 
Among all compared methods, only ROSEFusion~\cite{zhang2021rosefusion} and our approach successfully track all sequences, with our method consistently achieving superior results. Our lower errors demonstrate the robustness of our approach.

\begin{figure*}[tbh!]
\centering
\includegraphics[width=1.0\linewidth]{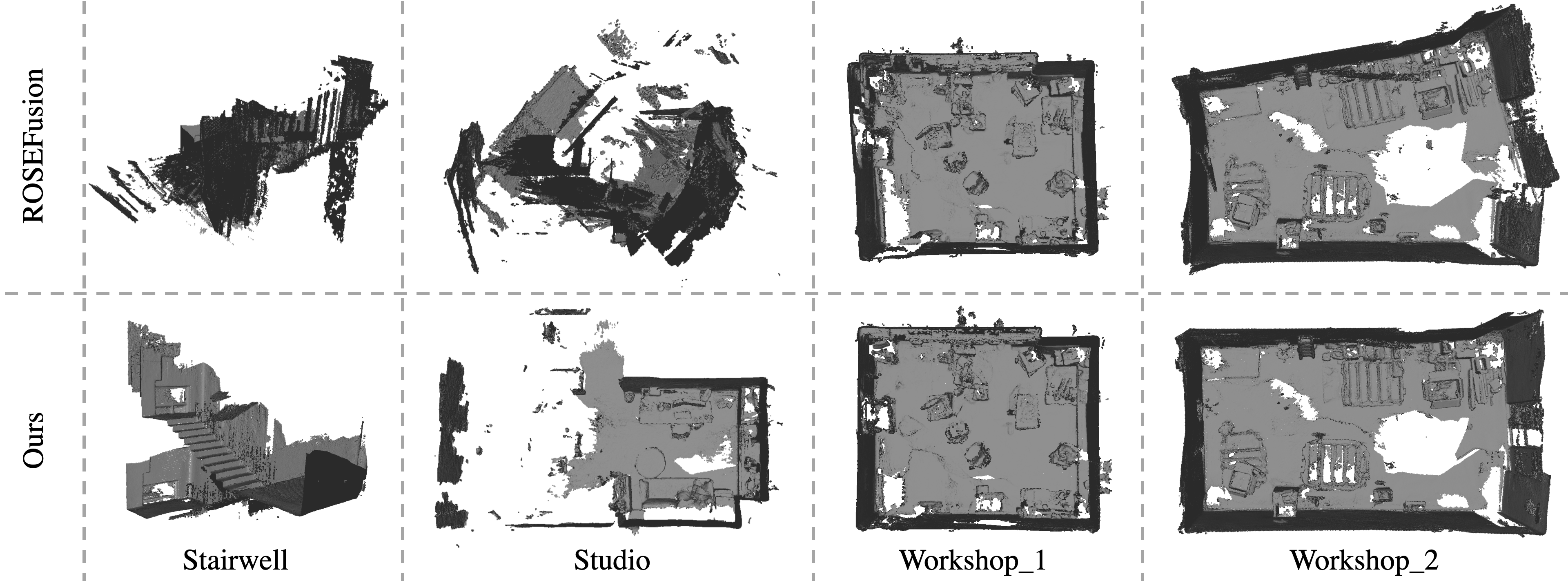}
\caption{
Visual comparison between ROSEFusion (the most robust competitor) and our system. We present dense reconstruction results from the four most challenging sequences from FastCaMo-Real. For each sequence, we drop 50\%-80\% of frames to mimic unstable motion. Our system performs only single-frame camera tracking without bundle adjustment or loop closure, yet the reconstructed layout demonstrates both robustness (no wrong registration) and accuracy (minimal drift). 
}
\label{fig:visual}
\end{figure*}

\begin{table}[tbh!]

\caption{Evaluation of reconstruction completeness and accuracy on FastCaMo-Real benchmark. We drop 50\%-80\% of frames for each sequence, making it more challenging.}

\centering

\resizebox{0.5\textwidth}{!}{


\begin{tabular}{l|cccc|c}
\toprule
Method & Stair. & Studio & Work.\_1 & Work.\_2 & Avg. \\
\toprule

\multirow{2}{*}{ROSEFusion~\cite{zhang2021rosefusion}} & - & - & 77.2\% & 52.2\% & - \\

& - & - & 3.2cm & 4.2cm & - \\

\midrule

\multirow{2}{*}{{Ours}} & \textbf{66.8\%} & \textbf{67.8\%} & \textbf{82.8\%} & \textbf{64.2\%} & \textbf{70.4\%} \\

& \textbf{2.8cm} & \textbf{2.6cm} & \textbf{3.1cm} & \textbf{3.5cm} & \textbf{3.0cm} \\

\bottomrule
\end{tabular}

}

\label{tab:fastcamo_real3}
\end{table}

\begin{figure}[tbh!]
\centering
\includegraphics[width=1.0\linewidth]{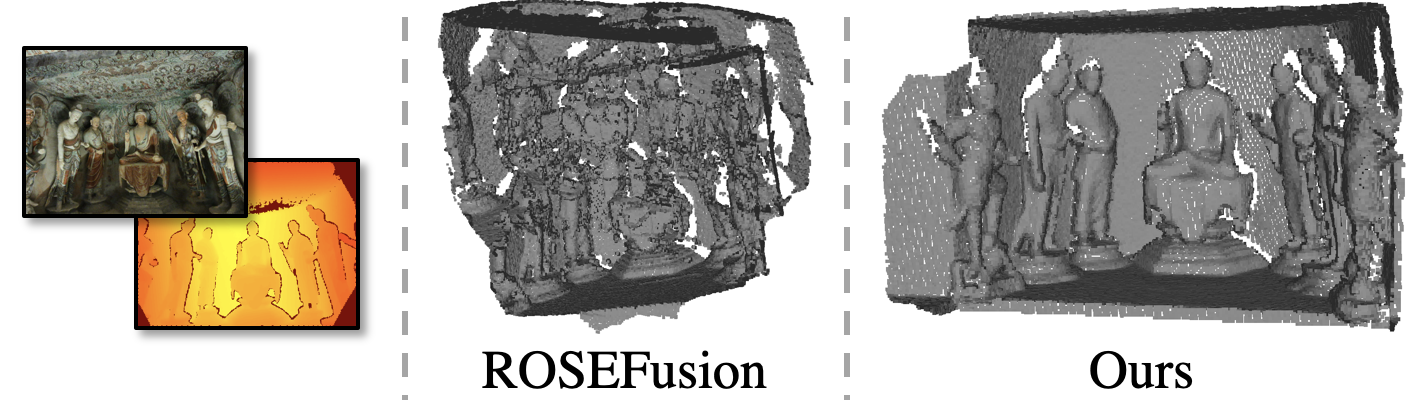}
\caption{
Despite being trained only on daily indoor scenes, our system generalizes well to novel environments such as cave sculptures. While ROSEFusion can handle moderately fast camera motion, sudden shaking can cause wrong registration and disrupt the reconstruction.}
\label{fig:mogaoku}
\end{figure}

\noindent
\textbf{FastCaMo-Synth~\cite{zhang2021rosefusion}.}
This benchmark contains 10 room-scale synthetic scenes. The RGB-D sequences are rendered from manually crafted fast camera motions. The average camera speed is 54.43 degrees/s and 1.68 m/s. The benchmark provides two sets of sequences for each scene: one with perfect color and depth images, and another with simulated motion blur and depth noise.
The camera tracking results are reported in Table~\ref{tab:fastcamo_raw} and \ref{tab:fastcamo}, respectively. 

Despite perfect RGB-D frames in the raw sequences, several systems struggle with tracking robustness. ElasticFusion~\cite{whelan2015elasticfusion} and NICE-SLAM~\cite{zhu2022nice} often produce incorrect registrations and messy reconstructions, while BundleFusion~\cite{dai2017bundlefusion} usually drops frames and loses tracking capability. HERO-SLAM~\cite{xin2024hero} successfully tracks all sequences using learning-based keypoint matching and bundle adjustment, but exhibits noticeable errors and operates $<$10 FPS. 
MIPS-Fusion~\cite{tang2023mips} tracks all sequences with randomized optimization, but its neural representation and hybrid optimization strategy compromise robustness compared to ROSEFusion~\cite{zhang2021rosefusion}'s pure randomized optimization approach. Our system outperforms all competing methods with superior robustness and accuracy, delivering the lowest camera tracking error.

When using noisy RGB-D frames, tracking accuracy degrades for all methods. Blur in color images specifically affects photometric losses in neural representations, while noisy depth data introduces errors in geometric-based optimization. These imperfect inputs also reduce the performance of our pose regression network and randomized optimization algorithm. Despite these challenges, our method still achieves the best overall accuracy compared to competing approaches. This further validates our robustness.

\noindent
\textbf{FastCaMo-Real~\cite{zhang2021rosefusion}.}
This benchmark contains 12 real-world scanning sequences featuring large-scale scenes. The benchmark doesn't provide ground-truth camera poses. Instead, it offers high-quality laser-scanned meshes as reference models. We align the scene reconstructions to these reference models using CloudCompare and then evaluate the completeness and accuracy of the 3D models. 
We calculate completeness by iterating through each point in the ground-truth model and determining the percentage of points successfully reconstructed (10cm threshold). Accuracy is measured as the average error of the reconstructed points. 

{
By combining pose regression and randomized optimization, our system provides both robustness and accuracy in camera pose estimation, ultimately yielding more favorable reconstruction results compared to competitors. The numerical results for the first 4 scenes are reported in Table~\ref{tab:fastcamo_real}. The methods not included in the table failed in most of the scenes. We can observe that ROSEFusion~\cite{zhang2021rosefusion} remains the most robust competitor in the literature. Overall, our reconstruction results achieve the best completeness and accuracy. 
}

{
To further validate our robustness, we make the remaining scenes more challenging by dropping some frames in the sequences to simulate more unstable RGB-D streams. 
As shown in Table~\ref{tab:fastcamo_real2}, our advantage over ROSEFusion~\cite{zhang2021rosefusion} becomes evident with 20\% of frames dropped. 
When dropping 50\%-80\% of frames,  ROSEFusion~\cite{zhang2021rosefusion} produces noticeable errors and fails in two scenes, while our method remains reliable. 
Figure~\ref{fig:visual} provides a visual comparison, and the numerical results are reported in Table~\ref{tab:fastcamo_real3}. 
}

\noindent
\textbf{Real-world applications.}
We captured several RGB-D videos in different environments and at various frame rates. We first visualize the reconstruction results from two representative sequences in Figure~\ref{fig:teaser}. The first sequence is at 5 FPS, featuring large viewpoint changes but less motion blur, while the second is at 30 FPS with fast in-place rotation. We compared our system with the strongest competitor, ROSEFusion~\cite{zhang2021rosefusion}. While ROSEFusion~\cite{zhang2021rosefusion} produces noticeable incorrect registrations, our method demonstrates robust performance. 
In Figure~\ref{fig:mogaoku}, we visualize the reconstruction results from a sequence of sculptures scanned in a cave with sudden camera shaking. Our clean reconstruction results demonstrate the robustness and generalizability of our approach, despite the network being trained only on daily indoor scenes.

\subsection{Analyses}

In this section, we showcase why both pose regression and randomized optimization are necessary components of our system. We report runtime statistics and also discuss limitations along with directions for future work.

\begin{figure}[tbh!]
\centering
\includegraphics[width=1.0\linewidth]{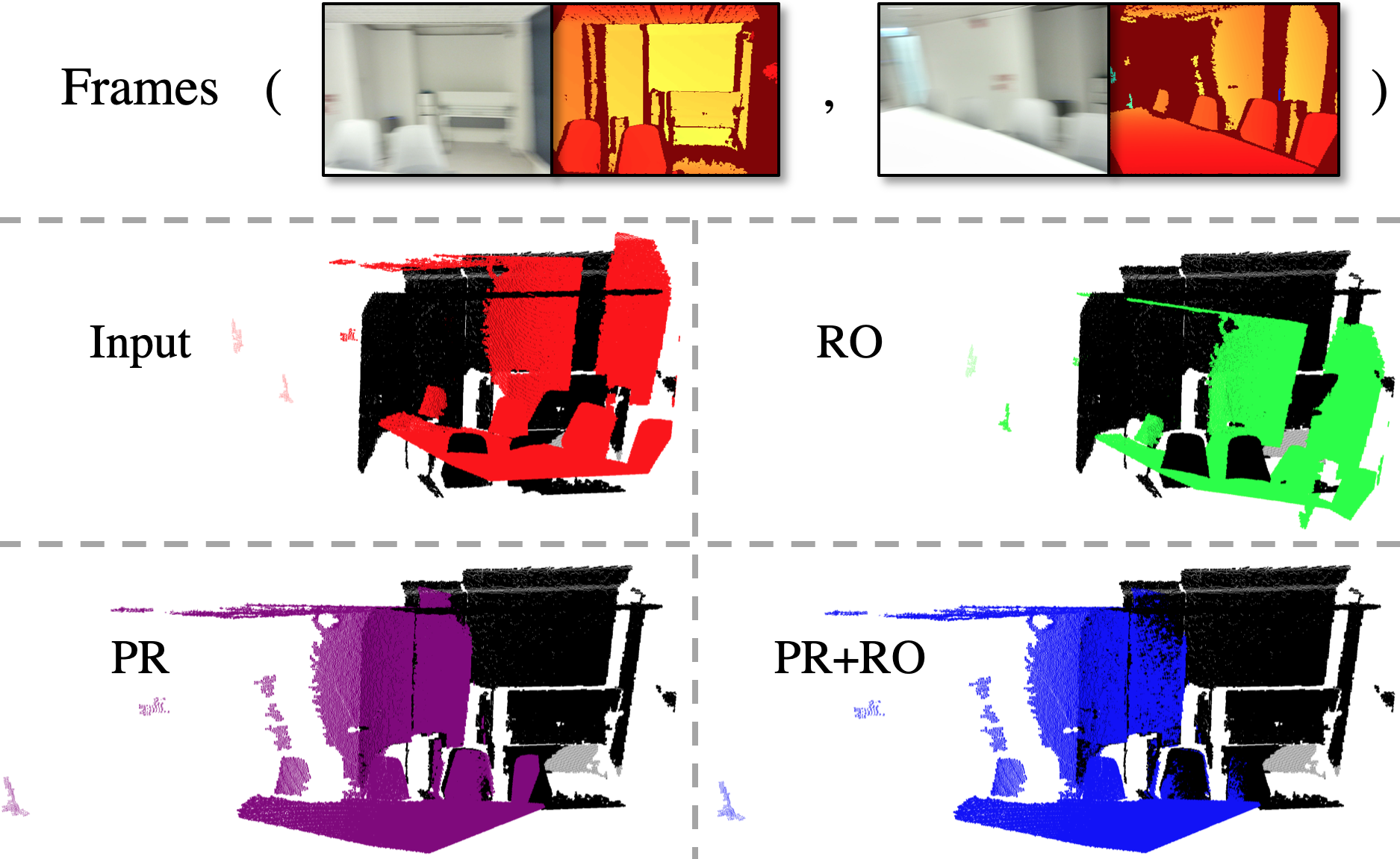}
\caption{Visual comparison with different methods. }
\label{fig:ablation}
\end{figure}

\noindent
\textbf{Ablation study.}
The key design of our system combines pose regression (PR) and randomized optimization (RO). To verify their effectiveness, we run experiments on FastCaMo-Synth (raw) and analyze the statistics of relative pose estimates. As shown in Table~\ref{tab:relpose}, PR alone lacks accuracy (higher minimum error), while RO alone lacks robustness (higher maximum error). Figure~\ref{fig:ablation} illustrates this trade-off: PR exhibits drift, and RO produces incorrect registration. Combining them achieves both robustness and accuracy.

\begin{table}[tbh!]

\caption{Relative pose errors with different methods.}

\centering

\resizebox{0.38\textwidth}{!}{


\begin{tabular}{l|c|c|c}
\toprule
Method & Median & Minimum & Maximum \\
\toprule

PR & \underline{0.26cm} & \underline{0.16mm} & \underline{2.60cm} \\

RO & 0.27cm & \textbf{0.06mm} & 5.55cm \\

Full (PR+RO) & \textbf{0.25cm} & \textbf{0.06mm} & \textbf{2.12cm} \\

\bottomrule
\end{tabular}

}

\label{tab:relpose}
\end{table}

\noindent
\textbf{Runtime statistics.}
{Our experiments are conducted on a server with an Intel® Xeon® Silver 4314 CPU (2.40GHz $\times 16$) and an NVIDIA RTX 4090 GPU.} 
In Table~\ref{tab:runtime}, we report our running statistics for reconstructing the scene Frl\_apartment\_2 from FastCaMo-Synth (noise) benchmark. We use a TSDF resolution of 2cm and a grid size of $500 \times 300 \times 750$ (sufficient to cover 150 m$^2$). Our pose regression network takes the main GPU memory consumption, while TSDF memory usage varies with resolution and grid size. Across all experiments, the total GPU memory consumption of our system remains under 10GB. Our network is purely feed-forward, achieving inference times under 20ms. Our randomized optimization runs in parallel using CUDA, requiring only a few milliseconds. 
Overall, our system can deliver real-time performance over 30 FPS.

\begin{table}[tbh!]

\caption{Resource consumption and running time.}

\centering

\resizebox{0.47\textwidth}{!}{


\begin{tabular}{cc|cc|c}
\toprule
\multicolumn{2}{c|}{Peak memory (GB)} & \multicolumn{2}{|c|}{Running time (ms)} & \multirow{2}{*}{Frames per second} \\

PR & TSDF+RO & PR & RO \\

\toprule

6.76 & 1.19 & $<$20 & $<$10 & $>$30 \\

\bottomrule
\end{tabular}

}

\label{tab:runtime}
\end{table}

\noindent
\textbf{Limitations.} 
A noticeable limitation of our system is its reliance on single-frame tracking without integrating bundle adjustment or loop closure. This lack of global optimization can lead to accumulated drift in very large scenes. In addition, our method typically fails when input frames lack features, such as completely blurred images or frames with no overlap due to extremely fast motion. These conditions make the registration problem ill-posed. Integrating IMU data could potentially solve this challenge. Addressing these limitations will be our focus in future work. 
\section{CONCLUSION}

In this paper, we present a robust and accurate dense scene reconstruction system. This is achieved through a simple combination of a camera pose regression network and a randomized optimization algorithm. Despite its simplicity, the system provides high-quality reconstruction results in real-time, regardless of camera motion stability. This makes it suitable for robot applications such as exploration and rescue scenarios that potentially involve unstable motions.

\noindent
\textbf{Acknowledgment.}
This work is supported by the Early
Career Scheme of the Research Grants Council (grant \# 27207224), the Hong Kong STEM Professorship program, and the JC STEM Lab of Autonomous Intelligent Systems funded by The Hong Kong Jockey Club Charities Trust.
We thank the editors and reviewers for their valuable suggestions and Yuzheng Liu, Jiazhao Zhang, and Shuzhe Wang for their help with experiments and proofreading.

\bibliographystyle{IEEEtran}
\bibliography{siyan}

\end{document}